\documentclass[12pt,a4paper,onecolumn,twoside,openright]{ranlp_bk}

\usepackage{epsf}        
\usepackage{epsfig}
\usepackage{makeidx}     
\makeindex               

\renewcommand{\contentsname}{CONTENTS}
\newcommand{\ranlptitle}[1]{    \chapter*{#1}
                                \setcounter{section}{0}
                                \setcounter{figure}{0}
                                \setcounter{table}{0}
                                \setcounter{equation}{0}
                                \setcounter{footnote}{0}   }

\renewcommand{\author}[1]{\vspace*{-2mm}
                          \begin{center}\textsc{#1}\end{center}}

\newcommand{\affiliation}[1]{\begin{center}\emph{#1}\end{center}
                             \vspace*{3mm}}

\newcommand{\ranlpaddr}[1]{}    

\newcommand{\runningheads}[2]{  
        \markboth{\small\uppercase{#1}}{\small\uppercase{#2}}}

\newcommand{\ranlpabstract}[1]{   \centerline{\textbf{Abstract}}\vspace*{-1mm}
                                  \begin{quote}\small #1 \end{quote} }

\renewcommand{\figurename}{Fig.} 

\newcommand{\ranlpfig}[4]{
  \begin{figure}[thb]
    \vspace*{-1ex}
    \hrule\vspace*{2ex minus 1ex}
    \begin{center}
      \mbox{
        \epsfxsize=#1\textwidth
        \epsfbox{#2}}
    \end{center}
    \vspace*{-1.5ex}
    \caption{#4}\label{#3}
    \hrule\vspace*{-2ex}\vspace*{0.5ex minus 0.4ex}
  \end{figure}}

\renewenvironment{thebibliography}{
        \vspace{1.7em}\small \centerline{\bf REFERENCES}\vspace*{1mm}
        \begin{list}{}
                {\setlength{\itemsep}{-0.6mm}
                 \setlength{\itemindent}{-7mm}
                 \setlength{\leftmargin}{7mm} 
                }               }{\end{list}}

\makeatletter
\long\def\@makefntext#1{\@setpar{\@@par\@tempdima \hsize 
  \advance\@tempdima-10pt\parshape \@ne 10pt \@tempdima}\par
  \parindent 1em\noindent \hbox to \z@{\hss$^{\@thefnmark}\ $}#1}
\makeatother

\renewcommand{\indexname}{Index of Subjects and Terms}

\newcommand{\hush}[1]{}             

\newlength{\spacing}
\newcommand{\nspace}[1]{\setlength{\baselineskip}{#1\spacing}}

\begin{document}
   \pagenumbering{arabic}
\ranlptitle{Part-of-Speech Tagging with Minimal Lexicalization}

\author{Virginia Savova$^*$ \& Leonid Peshkin$^{**}$}
\affiliation{$^*$Johns Hopkins University, \\
        $^{**}$Massachusetts Institute of Technology}
\runningheads{VIRGINIA SAVOVA \& LEONID PESHKIN}
       {POS TAGGING WITH MINIMAL LEXICALIZATION}

\newcommand{\spind}{SP$_{\mbox{ind}}$}
\newcommand{\spindz}{{\small\em SP}$_{\mbox{\em ind}}$}
\newcommand{\spcat}{{\small\em SP}$_{\mbox{cat}}$}
\newcommand{\spcatz}{{\small\em SP}$_{\mbox{\em cat}}$}
\newcommand{\spvalz}{{\small\em SP}$_{\mbox{\em val}}$}
\newcommand{\spval}{{\small\em SP}$_{\mbox{val}}$}
\newcommand{\spx}{{\small\em SP}$_{\mbox{\em x}}$}
\newcommand{\tcm}{{\small TCM}}
\newcommand{\SA}{{\small SA}}

\ranlpabstract{We use a Dynamic Bayesian Network ({\sc dbn}) to represent compactly a variety of sublexical and contextual features relevant to Part-of-Speech (PoS) tagging. The outcome is a flexible tagger (LegoTag) with state-of-the-art performance (3.6\% error on a benchmark corpus). We explore the effect of eliminating redundancy and radically reducing the size of feature vocabularies. We find that a small but linguistically motivated set of suffixes results in improved cross-corpora generalization. We also show that a minimal lexicon limited to function words is sufficient to ensure reasonable performance.}

\section{Part-of-Speech Tagging}\index{Part-of-Speech Tagging}

Many NLP applications are faced with the dilemma whether to use statistically extracted or expert-selected features\index{feature selection}. There are good arguments in support of either view. Statistical feature selection does not require extensive use of human domain knowledge, while feature sets chosen by experts are more economical and generalize better to novel data.

Most currently available PoS taggers\index{taggers}\index{PoS taggers} perform with a high degree of accuracy. However, it appears that the success in performance can be overwhelmingly attributed to an across-the-board lexicalization of the task. Indeed, Charniak, Hendrickson, Jacobson \& Perkowitz (1993) note that a simple strategy of picking the most likely tag for each word in a text leads to 90\% accuracy. If so, it is not surprising that taggers using vocabulary lists, with number of entries ranging from 20k to 45k, perform well. Even though a unigram model achieves an overall accuracy of 90\%, it relies heavily on lexical information and is next to useless on nonstandard texts that contain lots of domain-specific terminology. 

The lexicalization of the PoS tagging task comes at a price. Since word lists are assembled from the training corpus, they hamper generalization across corpora. In our experience, taggers trained on the Wall Street Journal ({\sc wsj})\index{Wall Street Journal} perform poorly on novel text such as email or newsgroup messages (a.k.a. Netlingo)\index{Netlingo}. At the same time, alternative training data are scarce and expensive to create. This paper explores an alternative to lexicalization. Using linguistic knowledge, we construct a minimalist tagger with a small but efficient feature set, which maintains a reasonable performance across corpora. 

A look at the previous work on this task reveals that the unigram model is at the core of even the most sophisticated taggers. The best-known rulebased tagger (Brill 1994) works in two stages: it assigns the most likely tag to each word in the text; then, it applies transformation rules of the form ``Replace tag X by tag Y in triggering environment Z''. The triggering environments span up to three sequential tokens in each direction and refer to words, tags or properties of words within the region. The Brill tagger achieves less than 3.5\% error on the Wall Street Journal ({\sc wsj}) corpus. However, its performance depends on a comprehensive vocabulary (70k words). 

Statistical tagging is a classic application of Markov Models ({\sc mm}s) \index{Markov Models}. Brants (2000) argues that second-order {\sc mm}s can also achieve state-of-the-art accuracy, provided they are supplemented by smoothing techniques and mechanisms to handle unknown words. TnT handles unknown words by estimating the tag probability given the suffix of the unknown word and its capitalization. The reported 3.3\% error for Trigrams 'n Tags (TnT) tagger on the {\sc wsj} (trained on $10^6$ words and tested on $10^4$) appears to be a result of overfitting. Indeed, this is the maximum performance obtained by training TnT until only 2.9\% of words are unknown in the test corpus. A simple examination of {\sc wsj} shows that such percentage of unknown words in the testing section (10\% of {\sc wsj} corpus) requires simply building a unreasonably large lexicon of nearly all (about 44k) words seen in the training section (90\% of {\sc wsj}), thus ignoring the danger of overfitting. Hidden {\sc mm}s ({\sc hmm}s) are trained on a dictionary with information about the possible PoS of words (Jelinek 1985; Kupiec 1992). This means {\sc hmm} taggers also rely heavily on lexical information.

Obviously, PoS tags depend on a variety of sublexical features, as well as on the likelihood of tag/tag and tag/word sequences. In general, all existing taggers have incorporated such information to some degree. The Conditional Random Fields ({\sc crf}) model (Lafferty, McCallum \& Pereira 2002) outperforms the {\sc hmm} tagger on unknown words by extensively relying on orthographic and morphological features. It checks whether the first character of a word is capitalized or numeric; it also registers the presence of a hyphen and morphologically relevant suffixes (-ed, -ly, -s, -ion, -tion, -ity, -ies). The authors note that {\sc crf}-based taggers are potentially flexible because they can be combined with feature induction algorithms. However, training is complex (AdaBoost~+~Forward-backward) and slow ($10^3$ iterations with optimized initial parameter vector; fails to converge with unbiased initial conditions). It is unclear what the relative contribution of features is in this model. 

The Maximum Entropy tagger (MaxEnt\index{Maximum entropy}, see Ratnaparkhi 1996) accounts for the joint distribution of PoS tags and features of a sentence with an exponential model. Its features are along the lines of the {\sc crf} model:

\begin{enumerate}
	\item {\bf Capitalization}: Does the token contain a capital letter?;
	\item {\bf Hyphenation}: Does the token contain a hyphen?;
	\item {\bf Numeric}: Does the token contain a number?;
	\item {\bf Prefix}: Frequent prefixes, up to 4 letters long;
	\item {\bf Suffix}: Frequent suffixes, up to 4 letters long;
\end{enumerate}

In addition, Ratnaparkhi uses lexical information on frequent words in the context of five words. The sizes of the current word, prefix, and suffix lists were 6458, 3602 and 2925, respectively. These are supplemented by special Previous Word vocabularies. Features frequently observed in a training corpus are selected from a candidate feature pool. The parameters of the model are estimated using the computationally intensive procedure of Generalized Iterative Scaling ({\sc gis})to maximize the conditional probability of the training set given the model. MaxEnt tagger has 3.4\% error rate. Our investigation examines how much lexical information can be recovered from sublexical features. In order to address these issues we reuse the feature set of MaxEnt in a new model, which we subsequently minimize with the help of linguistically motivated vocabularies. 

\section{PoS Tagging Bayesian Net}\index{Bayesian Net}

Our tagger combines the features suggested in the literature to date into a Dynamic Bayesian Network ({\sc dbn}). We briefly introduce the essential aspects of {\sc dbn}s here and refer the reader to a recent dissertation(Murphy 2002) for an excellent survey. A {\sc dbn} is a Bayesian network unwrapped in time, such that it can represent dependencies between variables at adjacent time slices. More formally, a {\sc dbn} consists of two models $B^0$ and $B^+$, where $B^0$ defines the initial distribution over the variables at time $0$, by specifying:

\begin{itemize}
\item set of variables $X_{1} , \ldots, X_{n}$;
\item directed acyclic graph over the variables;
\item for each variable $X_{i}$ a table specifying the conditional\\ probability of $X_{i}$ given its parents in the graph $Pr(X_{i}|Par\{X_i\})$.
\end{itemize}
The joint probability distribution over the initial state is: 
	\[\Pr(X_1,...,X_n)=\prod_1^n\Pr(X_i|Par\{X_i\}).\]
The transition model $B^+$ specifies the conditional probability distribution ({\sc cpd}) over the state at time $t$ given the state at time $t\!-\!1$. $B^+$ consists of: 
\begin {itemize}
\item directed acyclic graph over the variables $X_1,\ldots,X_n$ and their predecessors $X_1^-,\ldots,X_n^-$ --- roots of this graph;
\item conditional probability tables $\Pr(X_i|Par\{X_i\})$ for all $X_i$~(but not $X_i^-$).
\end {itemize}
The transition probability distribution is: 
\[\Pr(X_1,...,X_n\Big|X_1^-,...,X_n^{-})=\prod_1^n\Pr(X_i|Par\{X_i\}).\]
Between them, $B^0$ and $B^+$ define a probability distribution over the realizations of a system through time, which justifies calling these {\sc bn}s ``dynamic''. In our setting, the word's index in a sentence corresponds to time, while realizations of a system correspond to correctly tagged English sentences. Probabilistic reasoning about such system constitutes inference. 

Standard inference algorithms for {\sc dbn}s are similar to those for {\sc hmm}s. Note that, while the kind of {\sc dbn} we consider could be converted into an equivalent {\sc hmm}, that would render the inference intractable due to a huge resulting state space. In a {\sc dbn}, some of the variables will typically be observed, while others will be hidden. The typical inference task is to determine the probability distribution over the states of a hidden variable over time, given time series data of the observed variables. This is usually accomplished using the forwardbackward algorithm. Alternatively, we might obtain the most likely sequence of hidden variables using the Viterbi algorithm. These two kinds of inference yield resulting PoS tags. Note that there is no need to use ``beam search'' (cf. Brants 2000). 

Learning the parameters of a {\sc dbn} from data is generally accomplished using the EM algorithm. However, in our model, learning is equivalent to collecting statistics over cooccurrences of feature values and tags. This is implemented in {\sc gawk} scripts and takes minutes on the {\sc wsj} training corpus. Compare this to {\sc gis} or {\sc iis} (Improved Iterative Scaling) used by MaxEnt. In large {\sc dbn}s, exact inference algorithms are intractable, and so a variety of approximate methods has been developed. However, as we explain below, the number of hidden state variables in our model is small enough to allow exact algorithms to work. For the inference we use the standard algorithms, as implemented in the Bayesian network toolkit ({\sc bnt}, see Murphy 2002).

\ranlpfig{0.70}{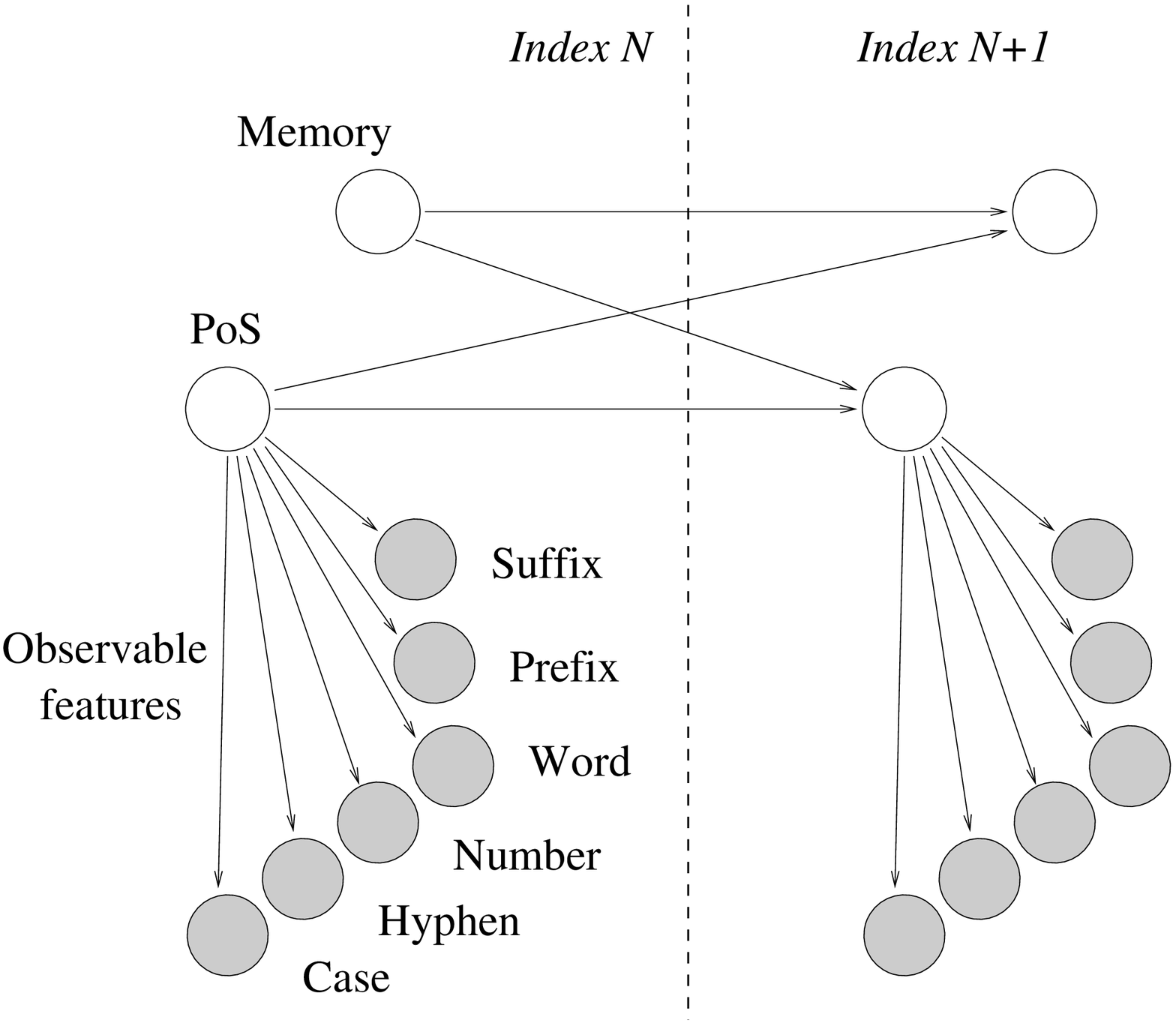}{savova:fig:1}{{\sc dbn} for PoS tagging.}

We base our original {\sc dbn} on the feature set of Ratnaparkhi's MaxEnt: the set of observable nodes in our network consists of the current word $W_i$, a set of binary variables $C_i$, $H_i$ and $N_i$ (for Capitalization, Hyphen and Number) and multivalued variables $P_i$ and $S_i$ (for Prefix and Suffix), where subscript i stands for position index. There are two hidden variables: $T_i$ and $M_i$ (PoS and Memory). Memory represents contextual information about the antepenultimate PoS tag. A special value of Memory (``Start'') indicates the beginning of the sentence. The PoS values are $45$ tags of the Penn Treebank tag set (Marcus, Kim, Marcinkiewicz, MacIntyre, Bies, Ferguson, Katz \& Schasberger 1994).

Figure~1 represents dependencies among the variables. Clearly, this model makes a few unrealistic assumptions about variable independence and Markov property of the sequence. Empirically this does not present a problem. For the discussion of these issues please see Bilmes (2003) who is using similar models for speech recognition. Thus, probability of a complete sequence of PoS tags $T_1 \ldots T_n$ is modeled as: \\
$\Pr(T_1 \ldots T_n) = \Pr(T_1)\times\Pr(F_1|T_1)\times\Pr(T_2|T_1, Start)\times\Pr(F_2|T_2)\times\Pr(M_2|T_1)\\
{\hskip 2.3 cm} \times\prod_{i-3}^{n-1}\Pr(T_i|T_{i-1}, M_{i-1})\times\Pr(M_i|T_{i-1}, M_{i-1})\times \Pr(F_i|T_i)\\
{\hskip 2.3cm} \times\Pr(T_n|T_{n-1}, M_{n-1})\times\Pr(F_n|T_n)$,\\
where $F_i$ is a set of features at index $i \in [1..n]$ and:
$$\Pr(F_i|T_i)\!=\!\Pr(S_i|T_i)\!\times\!\Pr(P_i\!|T_i)\!\times\Pr(W_i|T_i)\times\Pr(C_i|T_i)\!\times\Pr(H_i\!|T_i)\!\times\Pr(N_i\!|T_i)\;.$$ 
These probabilities are directly estimated from the training corpus. 

\section{Experiments and Results}

We use sections 0-22 of {\sc wsj} for training and sections 23, 24 as a final test set. The same split of the data was used in recent publications (Toutanova \& Manning 2002; Lafferty, McCallum \& Pereira 2001) that report relatively high performance on out-of-vocabulary (OoV) items. The test sections contain 4792 sentences out of about 55600 total sentences in {\sc wsj} corpus. The average length of a sentence is 23 tokens. The Brown corpus is another part of UPenn TreeBank dataset, which is of a similar size to {\sc wsj} (1016277 tokens) but quite different in style and nature. The Brown corpus has substantially richer lexicon and was chosen by us to test the performance on novel text.

We begin our experiments by combining the original MaxEnt feature set into a {\sc dbn} we call LegoTag to emphasize its compositional nature. The initial network achieves 3.6\% error (see Table~1) and closely matches that of MaxEnt (3.4\%).
\begin{table}[htb]
\begin{small}
 \begin{center}     
  \begin{tabular}{|l|l|l|l|l|}		
  \hline
  \textbf{Memory}& {\textbf{Features}}&\multicolumn{3}{|l|}{\textbf{Error}}\\\cline{3-5}
  \textbf{(of values)}& {}&\textbf{Ave}&\textbf{OoV}&\textbf{Sen}\\ \hline
		{Clustered (5)}&Unfactored&$4.4$&$13.0$&$58.5$\\ \hline
		{Clustered (5)}&Factored&$3.9$&$10.8$&$55.8$\\ \hline
		{Full (45)}&Factored&$3.6$&$9.4$&$51.7$\\ \hline
  \end{tabular}
 \end{center}
\end{small}
\caption{Results for Full LegoTag on {\sc wsj}.}
\label{savova:table1}
\end{table}
Our first step is to reduce the complexity of our tagger because performing inference on the {\sc dbn} containing a conditional probability table of $45^3$ elements for Memory variable is cumbersome. At the cost of minor deterioration in performance (3.9\%, see Table~1), we compress the representation by clustering Memory values that predict similar distribution over Current tag values. The clustering method is based on Euclidian distance between $45^2$ dimensional probability vectors $\Pr(T_i|T_{i-1})$. We perform agglomerative clustering, minimizing the sparseness of clusters (by assigning a given point to the cluster whose farthest point it is closest to). As a result of clustering, the number of Memory values is reduced nine times. Consequently, the conditional probability table of Memory and PoS become manageable. 

As a second step to simplification of the network, we eliminate feature redundancy. We leave only the lowercase form of each word, prefix and suffix in the respective vocabulary; remove numbers and hyphens from the vocabulary, and use prefix, suffix and hyphen information only if the token is not in the lexicon. The size of the factored vocabularies for word, prefix and suffix is 5705, 2232 and 2420 respectively (a reduction of 12\%, 38\% and 17\%). Comparing the performance of LegoTag with factored and unfactored features clearly indicates that factoring pays off (Table~1). Factored LegoTag is better on unknown words and at the sentence level, as well as overall. In addition, factoring simplifies the tagger by reducing the number of feature values.
We report three kinds of results: overall error, error on unknown words (OoV), and per sentence error . Our first result (Table~2) shows the performance of our network without the variable Word. 
\begin{table}[htb]
\begin{small}
 \begin{center}     
  \begin{tabular}{|l|l|l|l|l|l|l|l|l|}		
  \hline
  \multicolumn{6}{|l|}{\textbf{Type of LegoTag}}&\multicolumn{3}{|l|}{\textbf{Error}}\\\hline
  \textbf{H}& \textbf{N}&\textbf{C}&\textbf{P}& \textbf{S}&\textbf{W}&\textbf{Ave}&\textbf{OoV}&\textbf{Sen}\\\hline
  \textbf{$+$}& \textbf{$+$}&\textbf{$+$}&\textbf{$+$}& \textbf{$+$}&\textbf{$-$}&$11.3$&$11.3$&$84.0$\\\hline
  \textbf{$-$}& \textbf{$-$}&\textbf{$+$}&\textbf{$-$}& \textbf{$-$}&\textbf{$+$}&$6.1$&$30.6$&$69.0$\\\hline
  \textbf{$-$}& \textbf{$-$}&\textbf{$-$}&\textbf{$-$}& \textbf{$-$}&\textbf{$+$}&$9.3$&$47.6$&$77.7$\\\hline
  \end{tabular}
 \end{center}
\end{small}
\caption{Results of de-lexicalized and fully lexicalized LegoTag for {\sc wsj}.}
\label{savova:table2}
\end{table}
Even when all words in the text are unknown, sublexical features carry enough information to tag almost 89\% of the corpus. 

Next, we test two degenerate variants: one which relies only on lexical information, and another which relies on lexical information plus capitalization. Lexical information alone does very poorly on unknown words, which comes to show that context alone is not enough to uncover the correct PoS.

We now turn to the issue of using the morphological cues in PoS tagging and create a linguistically ``smart'' network (Smart LegoTag), whose vocabularies contain a collection of function words, and linguistically relevant prefixes and suffixes assembled from preparatory materials for the English language section of college entrance examination (Scholastic Aptitude Test). The vocabularies are very small: 315, 100, and 72, respectively. The percentage of unknown words depends on vocabulary size. For the large lexicon of LegoTag it is less than 12\%, while for the Smart LegoTag (whose lexicon contains only function words which are few but very frequent), it is around 50\%. In addition, two hybrid networks are created by crossing the suffix set and word lexicon of the Full LegoTag and Smart LegoTag. 

The results for the Smart LegoTag, as well as for the Hybrid LegoTags are presented in Table~3. They suggest that nonlexical information is sufficient to assure a stable, albeit not stellar, performance across corpora. Smart LegoTag was trained on {\sc wsj} and tested on both {\sc wsj} and Brown corpora with very similar results. The sentence accuracy is generally lower for the Brown corpus than for the {\sc wsj} corpus, due to the difference in average length. The Hybrid LegoTag with big suffix set and small word lexicon was a little improvement over Smart LegoTag alone. Notably, however, it is better on unknown words than Full LegoTag on the Brown corpus. 

The best performance across corpora was registered by the second Hybrid LegoTag (with big word lexicon and small suffix set). This is a very interesting result indicating that the nonlinguistically relevant suffixes in the big lexicon contain a lot of idiosyncratic information about the {\sc wsj} corpus and are harmful to performance on different corpora.
\begin{table}[htb]
\begin{small}
 \begin{center}
 \begin{tabular}{|l|l|l|l|l|l|l|l|l|l|}	
  \hline
\multicolumn{2}{|l|}{\textbf{LegoTag}}&\multicolumn{6}{|l|}{\textbf{Error}}&\textbf{Unkn}&\textbf{Unkn}\\\cline {3-8} 
\multicolumn{2}{|l|}{\textbf{Ftrs}}&\multicolumn{3}{|l|}{\textbf WSJ}&\multicolumn{3}{|l|}{\textbf{Brown}}&{\textbf{Wrd \%}}&{\textbf{Wrd \%}}\\\cline {1-8} 
\textbf{Word}&\textbf{Suff}&\textbf{Ave}&\textbf{OoV}&\textbf{Sen} &\textbf{Ave}&\textbf{OoV}&\textbf{Sen}&{\textbf WSJ}&{\textbf Brown}\\\hline
$5705$&$2420$&$3.9$&$10.0$&$55.4$&$10.1$&$23.4$&$67.9$&$11.6$&$15.4$\\\cline {1-8}
$5705$&$72$&$4.4$&$14.0$&$58.7$&$7.7$&$21.9$&$69.3$&{}&{}\\\hline
$315$&$2420$&$6.4$&$10.5$&$70.3$&$10.1$&$17.8$&$76.7$&$49.2$&$40.8$\\\cline{1-8} 
$315$&$72$&$9.6$&$17.1$&$82.2$&$11.4$&$22.3$&$82.9$&{}&{}\\\hline
   \end{tabular}
 \end{center}
\end{small}
\caption{Results for Smart and Hybrid LegoTags.}
\label{savova:table3}
\end{table}
Full LegoTag and Smart LegoTag encounter qualitatively similar difficulties. Since function words are part of the lexicon of both networks, there is no significant change in the success rate over function words. The biggest source of error for both is the noun/adjective (NN/JJ) pair (19.3\% of the total error for Smart LegoTag; 21.4\% of the total error for Full LegoTag). By and large, both networks accurately classify the proper nouns, while mislabeling adverbs as prepositions and vice versa. The latter mistake is probably due to inconsistency within the corpus (see Ratnaparkhi 1996 for discussion). One place where the two networks differ qualitatively is in their treatment of verbs. Smart LegoTag often mistakes bare verb forms for nouns. This is likely due to the fact that a phrase involving ``to'' and a following word can be interpreted either analogously to ``to mom'' (to + NN) or analogously to ``to go'' (to + VB) in the absence of lexical information. Similar types of contexts could account for the overall increased number of confusions of verb forms with nouns with Smart LegoTag. On the other hand, Smart LegoTag is much better at separating bare verb forms (VB) from present tense verbs (VBP) because it does not rely on lexical information that is potentially confusing since both forms are identical. However, it often fails to differentiate present verbs (VBP) from past tense verbs (VBD), presumably because it does not recognize frequent irregular forms. Adding irregular verbs to the lexicon may be a way of improving Smart LegoTag.

\section{Conclusion}

 {\sc dbn}s provide an elegant solution to PoS tagging. They allow flexibility in selecting the relevant features, representing dependencies and reducing the number of parameters in a principled way. Our experiments with a {\sc dbn} tagger underline the importance of selecting an efficient feature set. Eliminating redundancies in the feature vocabularies improves performance. Furthermore, reducing lexicalization leads to a higher capacity for generalization. Delexicalized taggers make fewer errors on unknown words, which naturally results in more robust success rate across corpora. 

The relevance of a given feature to PoS tagging varies across languages. For example, languages with rich morphology would call for a greater reliance on suffix/prefix information. Spelling conventions may increase the role of the Capitalization feature (e.g. German). In the future, we hope to develop methods for automatic induction of efficient feature sets and adapt the {\sc dbn} tagger to other languages.

\end{document}